\documentclass{article}

\usepackage{microtype}
\usepackage{graphicx}
\usepackage{subfigure}
\usepackage{booktabs} 

\usepackage{hyperref}

\usepackage[accepted]{icml2025}

\usepackage{amsmath}
\usepackage{amssymb}
\usepackage{mathtools}
\usepackage{amsthm}
\usepackage{svg}
\write18{}

\usepackage[capitalize,noabbrev]{cleveref}

\theoremstyle{plain}

\theoremstyle{definition}

\theoremstyle{remark}

\usepackage[textsize=tiny]{todonotes}
\usepackage{multirow}
\usepackage{tabularx}

\icmltitlerunning{Self-Evaluation for Job-Shop Scheduling}

\begin{document}

\twocolumn[
\icmltitle{Self-Evaluation for Job-Shop Scheduling}



\icmlsetsymbol{equal}{*}

\begin{icmlauthorlist}
\icmlauthor{Imanol Echeverria}{equal,tec,ehu}
\icmlauthor{Maialen Murua}{tec}
\icmlauthor{Roberto Santana}{ehu}
\end{icmlauthorlist}

\icmlaffiliation{ehu}{Computer Science and Artificial Intelligence Department, University of the Basque Country, San Sebastian, Spain}
\icmlaffiliation{tec}{TECNALIA, Basque Research and Technology Alliance (BRTA), San Sebastian, Spain}

\icmlcorrespondingauthor{Imanol Echeverria}{imanol.echeverria@tecnalia.com}

\icmlkeywords{Self-Evaluation, Job-shop scheduling, Graph neural networks}

\vskip 0.3in
]



\printAffiliationsAndNotice{}  

\begin{abstract}
Combinatorial optimization problems, such as scheduling and route planning, are crucial in various industries but are computationally intractable due to their \emph{NP-hard} nature. Neural Combinatorial Optimization methods leverage machine learning to address these challenges but often depend on sequential decision-making, which is prone to error accumulation as small mistakes propagate throughout the process. Inspired by self-evaluation techniques in Large Language Models, we propose a novel framework that generates and evaluates subsets of assignments, moving beyond traditional stepwise approaches. Applied to the Job-Shop Scheduling Problem, our method integrates a heterogeneous graph neural network with a Transformer to build a policy model and a self-evaluation function. Experimental validation on challenging, well-known benchmarks demonstrates the effectiveness of our approach, surpassing state-of-the-art methods.
\end{abstract}

\section{Introduction}

Combinatorial optimization problems (COPs) are fundamental to tasks such as scheduling, resource allocation, and route planning, influencing key decisions across various industries. Given their inherent complexity, most COPs are \emph{NP-hard} \cite{papadimitriou1998}, making exact methods impractical for large-scale or real-time decision-making. Traditionally, specialized heuristics—often grounded in domain expertise—have been employed to provide feasible solutions under tight time constraints \cite{blum2003metaheuristics}.

\emph{Neural Combinatorial Optimization} (NCO) \cite{vinyals2015pointer, bello2017neural} has emerged as a data-driven framework for deriving heuristics by leveraging recurring patterns in problem instances. Most NCO approaches adopt a \emph{constructive} viewpoint, treating solution building as a sequential decision-making process naturally framed by a Markov Decision Process (MDP). This formulation allows for training policies that iteratively determine the next choice in a sequence of decisions, ultimately arriving at a complete solution. Nevertheless, constructing solutions through a sequence of individual choices can pose significant challenges, as each local decision can limit or skew subsequent options, making it increasingly difficult to ensure a high-quality final outcome.

A related and rapidly evolving area is that of Large Language Models (LLMs), which face the challenge of generating coherent sequences. One promising direction in LLM research is \emph{self-evaluation}, where an auxiliary function assesses outputs, identifies weaknesses, and guides the refinement of subsequent steps \cite{kadavath2022,xie2024self}. Inspired by these advances, we propose applying self-evaluation principles to COP decoding. However, this design significantly departs from the typical approach in LLMs, as the constraints involved in constructing the optimal solution of a combinatorial problem are of a fundamentally different nature.

The problem we aim to address is generating subsequences of actions closer to the optimal solution, a core challenge in many COPs. Conventional methods, which produce one action at a time, often fail to generate coherent and optimal subsequences, as they do not directly evaluate the quality of action subsets using a dedicated evaluation mechanism. To address this, our framework introduces a novel mechanism that evaluates subsets of actions collectively at each step. By jointly considering multiple actions, the approach improves entire subsequences, enhancing solution quality compared to traditional stepwise methods. This shift from isolated moves to collective evaluations enables a richer and more effective decision-making process.

\begin{figure}[h] \centering \includegraphics[width=1\linewidth]{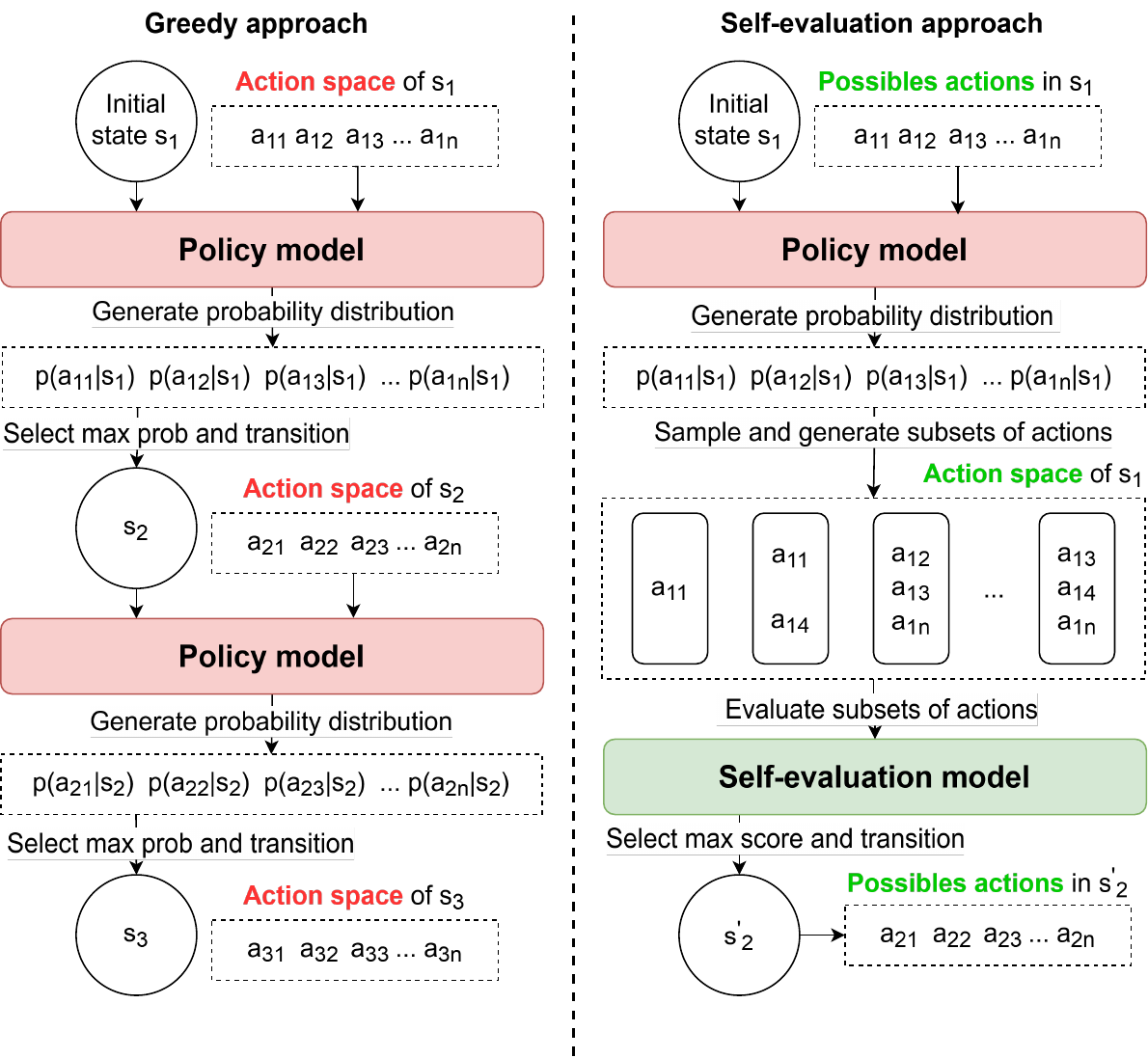} \caption{Comparison between a greedy approach and our self-evaluation framework.} \label{fig:selfevaluationsimple} \end{figure}

Our framework integrates two complementary models: a policy model and a self-evaluation model. The policy model assigns probabilities to possible actions using supervised learning, sampling sets of actions that are then scored by the self-evaluation model based on their overall quality. This allows for selecting the most promising subsequences during inference and redefines COPs as MDPs, transitioning from single-action spaces to subsequence-based spaces. Figure \ref{fig:selfevaluationsimple} illustrates how our approach evaluates entire subsets of actions jointly, unlike traditional greedy methods that assess actions individually. For instance, in routing problems, actions might represent cities to visit, while in scheduling, they could denote operation assignments to machines. Despite introducing an additional model, our subsequence-based design reduces the overall number of inference steps, making the framework efficient while better capturing dependencies between actions.

The neural network architecture of our approach follows a multi-model design for both the policy and self-evaluation components. The policy model combines a heterogeneous graph neural network (HGNN) with a Transformer to generate the probability distribution of the actions. The HGNN extracts structural information from the problem, leveraging the fact that many COPs can be represented as graphs, with information embedded in nodes, edges, or both. This makes our approach general and adaptable to a wide range of COPs. The self-evaluation function also uses the embeddings generated by the HGNN and is implemented as a separate Transformer-based module. This function outputs a score between 0 and 1, quantifying the quality of candidate action sets.

In this paper, we focus on scheduling—a more complex and less explored domain compared to other combinatorial optimization settings like routing. Specifically, we address the Job-Shop Scheduling Problem (JSSP), one of the most challenging problems in scheduling. We evaluated our approach on two well-known and challenging public JSSP benchmark datasets, comprising a total of 160 instances. The first dataset, the Taillard benchmark \cite{taillard1993benchmarks}, served as a reference for generating synthetic training instances. Specifically, we followed the same method used to create the benchmark's instances, ensuring they shared the same distribution, but we restricted the training instances to smaller problem sizes. This setup enables a controlled evaluation of the model’s ability to generalize to larger, unseen instances within the same distribution. In contrast, the second dataset, the Demirkol benchmark \cite{demirkol1998benchmarks}, features a distinct distribution, providing a more challenging test of the model's capacity to generalize to diverse and previously unseen scenarios. These benchmarks have been widely used in the literature to validate state-of-the-art approaches \cite{pirnay2024self, corsini2024self}.

Our experimental results demonstrate significant optimization improvements over state-of-the-art methods, with substantial gains on the dataset with a different distribution. It surpasses recent deep learning approaches and outperforms Google's optimization-focused OR-Tools CP-SAT solver on larger instances, even though the latter incurs significantly higher computational costs. These results emphasize the effectiveness of our framework in tackling diverse and unseen problem instances. While this work focuses on the JSSP, the proposed methodology holds potential for broader applications to other combinatorial optimization problems, which will be explored in future research.

The contributions of the paper can be summarized as follows:

\begin{itemize}
    \item A novel framework that evaluates subsets of actions collectively, refining subsequences to improve solution quality compared to traditional stepwise approaches.
    \item An integrated model design featuring an HGNN and Transformer for the policy model, coupled with a Transformer-based self-evaluation function for assessing action subsets.
    \item A redefinition of COPs as MDPs, transitioning from single-action spaces to subsequence-based action spaces for richer decision-making.
    \item Extensive evaluation on JSSP benchmarks, demonstrating significant performance gains over state-of-the-art methods, including superior generalization to diverse and unseen problem instances.
\end{itemize}

\section{Related Work}
\label{sec:state_of_the_art}

An early breakthrough in NCO leveraged supervised learning with recurrent neural networks, applied to COPs such as the Traveling Salesman Problem (TSP) \citep{vinyals2015pointer}. Although exact solvers are limited to small instances, they provided high-quality data to train neural networks capable of generalizing to larger problems, demonstrating the potential of learning-based methods for COPs.

Subsequent work, particularly in routing problems, shifted towards deep reinforcement learning (DRL) to train policies \cite{kool2018attention, kwon2020pomo}. These methods introduced various improvements in neural network architectures and strategies to exploit symmetries inherent in routing problems. Typically, they followed a \emph{constructive} approach, incrementally building solutions by selecting the next element—such as the next city in the TSP—until the solution was complete.

Building on the successes in routing, scheduling problems have also been addressed predominantly with deep reinforcement learning (DRL), incorporating various adjustments in policy training and state modeling \cite{song2022flexible, wang2023flexible}. In addition to recurrent neural networks and transformers \cite{tassel2023end, pirnay2024self}, commonly used in routing problems, scheduling methods leverage architectures specifically designed to handle the unique challenges of these problems. Unlike routing, scheduling involves more complex entities—such as operations, jobs, and machines—where HGNNs are often utilized \cite{song2022flexible}. HGNNs effectively represent multiple node and edge types, capturing the intricate relationships inherent in scheduling problems. However, most scheduling and routing methods adopt a \emph{constructive} approach, where operations are sequentially assigned, and policies are trained to make one assignment at a time. This severely limits their ability to capture broader dependencies between assignments, leading to suboptimal solutions. Additionally, research on hybridizing different architectures remains limited, as current approaches tend to exclusively rely on either recurrent neural networks or transformers, instead of combining their strengths to better handle the complexity of scheduling problems.

Although less common, \emph{neural improvement methods} provide an alternative to constructive approaches by refining an existing solution rather than building it step by step \cite{garmendia2023neural}. For example, in the JSSP \cite{zhang2024deep}, these methods learn policies—often via DRL variants—that rearrange operations pair by pair to optimize the execution sequence. However, this approach, where the action space is a set rather than a subsequence, faces similar challenges to constructive methods—namely, the difficulty of building subsequences of changes that are closer to the optimal solution.

Recently, researchers have begun exploring alternatives to standard DRL, in favor of self-supervised strategies \cite{corsini2024self, pirnay2024self}, a return to supervised methods \cite{drakulic2024bq}, or offline RL \cite{echeverria2024offline}, to overcome DRL’s limitations, most notably the extensive interactions required for exploration. Some approaches have started assigning more than one action simultaneously, such as the method in \citep{tassel2023end}, which combines policy gradient methods with imitation learning, and the approach in \citep{echeverria2024multi}, which relies on supervised learning. However, these methods lack a self-evaluation mechanism and do not move beyond the standard MDP framework. Actions are assigned solely based on the probabilities provided by the policy model, without evaluating how close the generated sequence is to the optimal solution. This can result in action sets lacking internal coherence, as they are not directly evaluated using a dedicated evaluation function.

The main contribution of this paper is to depart from generating solutions action by action and, instead, produce and evaluate subsequences collectively using a \emph{self-evaluation} mechanism inspired by the success of LLMs. This approach avoids the traditional stepwise paradigm of predicting a single move without supervision. Additionally, we integrate HGNNs with Transformers, both to generate the policy and to define the \emph{self-evaluation} function.

\section{Preliminaries}
\label{sec:preliminaries}

\textbf{Job Shop Scheduling Problem.} The JSSP is a classical combinatorial optimization problem defined as follows. Let $J = \{J_1, J_2, \dots, J_n\}$ represent a set of $n$ jobs, and let $M = \{M_1, M_2, \dots, M_m\}$ denote a set of $m$ machines. Each job $J_i$ consists of a sequence of $k_i$ operations, $\{O_{i1}, O_{i2}, \dots, O_{ik_i}\}$, where each operation $O_{ij}$ requires exclusive use of a specific machine $M_k \in M$ for a fixed processing time $p_{ij} > 0$. Operations within a job must follow a predefined order, meaning that $O_{ij}$ must complete before $O_{i(j+1)}$ begins for all $j = 1, \dots, k_i - 1$. Additionally, each machine $M_k$ can process at most one operation at a time, and once an operation starts on its assigned machine, it must run without interruption until completion.

The objective is to determine a feasible schedule $\mathcal{S}$ that minimizes the \emph{makespan}, $C_{\max}$, defined as the total time required to complete all jobs:
\begin{align}
C_{\max} = \max_{i=1, \dots, n} C_i,
\end{align}
where $C_i$ is the completion time of job $J_i$.

\textbf{Self-Evaluation in Large Language Models.} Breaking down complex tasks into intermediate steps, often referred to as reasoning chains, has proven highly effective for enhancing the performance of LLMs on multi-step problems \cite{brown2020language, wei2022chain}. By structuring tasks into logical sub-components, these models can process and solve problems more systematically. However, as reasoning chains grow longer, errors can accumulate across intermediate steps, significantly reducing the accuracy of the final outcomes \cite{chen2024self}.

Self-evaluation provides a mechanism to address these challenges by enabling models to critique and refine their intermediate reasoning. This approach involves evaluating the correctness and coherence of each step in the reasoning chain, allowing for dynamic adjustments and reducing the risk of error propagation \cite{madaan2024self}. By integrating explicit feedback into the reasoning process, self-evaluation ensures more consistent and robust outputs, particularly in tasks requiring logical precision and multi-step decision-making.

\section{Method}
\label{sec:method}

In this section, we present our contribution, which introduces a novel approach to applying the concept of \emph{self-evaluation} to scheduling problems. We begin by explaining how the problem is defined as a Markov Process and then detail the generation of the policy and \emph{self-evaluation} models. 




\subsection{JSSP as a Markov Process}

The JSSP is modeled as a Markov Process, with states representing scheduling progress and transitions corresponding to sets of job-machine assignments. Below, we outline the state space, action space, and transition function.

\textbf{State Space.} The state space is modeled as a heterogeneous graph $\mathcal{G}_t = (\mathcal{V}_t, \mathcal{E}_t)$, following the approach described in \cite{song2022flexible}. A simplified version, as proposed in \cite{echeverria2024multi}, is used in this work. At each timestep $t$, $\mathcal{G}_t$ includes three types of nodes: operations, jobs, and machines, as well as three distinct types of edges (directed and undirected) that capture relationships such as precedence constraints between operations, which operations belong to a specific job, and which operations can be executed by specific machines. Detailed information about the features of the different nodes and edges is provided in Appendix \ref{app:features}.

\textbf{Action Space.} At each timestep $t$, the set of feasible job-machine pairs is denoted as $\mathcal{JM}_t$. The action space is defined as the power set of all pairs $(j, m) \in \mathcal{JM}_t$ such that no machine is assigned more than one job simultaneously:
\begin{align}
\mathcal{A}_t = \{A \subseteq \mathcal{JM}_t \mid \forall (j, m), (j', m) \in A, j = j' \}.
\end{align}
This formulation allows multiple job-machine assignments to be made at once, provided they respect machine capacity constraints. This represents a significant shift from prior work \cite{echeverria2024multi} and many other scheduling \cite{wang2023flexible, tassel2023end} or routing \cite{kwon2020pomo, drakulic2024bq} approaches, where the action space is typically defined as a single assignment set. In scheduling, prior to this work, the action space would generally be defined as $\mathcal{JM}_t$.

\textbf{Transition Function.} The state transitions are determined by the selected assignments. Specifically, at timestep $t$, the selected action $\mathcal{A} \in \mathcal{A}_t$ updates the graph $\mathcal{G}_t$ by modifying the attributes of nodes and edges to reflect the assignments made. Following the work in \cite{ho2024residual, echeverria2025diverse}, completed operations are removed to simplify the state space.

In this work, we do not explicitly define a reward function, as both the policy and self-evaluation models are trained using supervised learning on optimal solutions. An example of how the action set is defined and state transitions are performed is provided in Appendix \ref{app:example}.

\subsection{Self-Evaluation}

We refer to our method as SEVAL (Self-evaluation), which introduces a novel mechanism for evaluating subsets of actions collectively, enhancing solution quality compared to traditional step-by-step approaches. SEVAL relies on two complementary models: a policy model that proposes candidate assignments, and a self-evaluation model that scores these sets of assignments, enabling the selection of the most promising ones during inference. In this section, we describe the dataset generation process, the training procedure for both models, and the inference mechanism.

\subsubsection{Dataset Generation}

To train both models, a supervised learning approach is employed, requiring a dataset. Some methods, like those in \cite{kool2019attention, drakulic2024bq}, rely solely on expert trajectories, which can limit data diversity and hinder performance in complex scenarios. The dataset $\mathcal{D}$ is constructed by solving small-scale JSSP instances with an efficient solver, transforming each solution into sequences of states and actions:
\[
\mathcal{D} = \{(s, \mathcal{A}) \mid s \in \mathcal{S}, \mathcal{A} \subseteq \mathcal{A}_t \},
\]
where $s$ is a heterogeneous graph of the scheduling state, and $\mathcal{A}$ represents feasible job-machine assignments. 

A key issue with supervised learning is poor generalization \cite{ross2010efficient}, especially when policies encounter underrepresented states during testing. The value of adding suboptimal trajectories to enhance offline reinforcement learning was emphasized in \cite{kumar2021should, andres2025using}. Appendix \ref{app:dataset} details how perturbing optimal instances creates a more diverse training set, improving generalization.

 \begin{figure*}[h] \centering \includegraphics[width=1\linewidth]{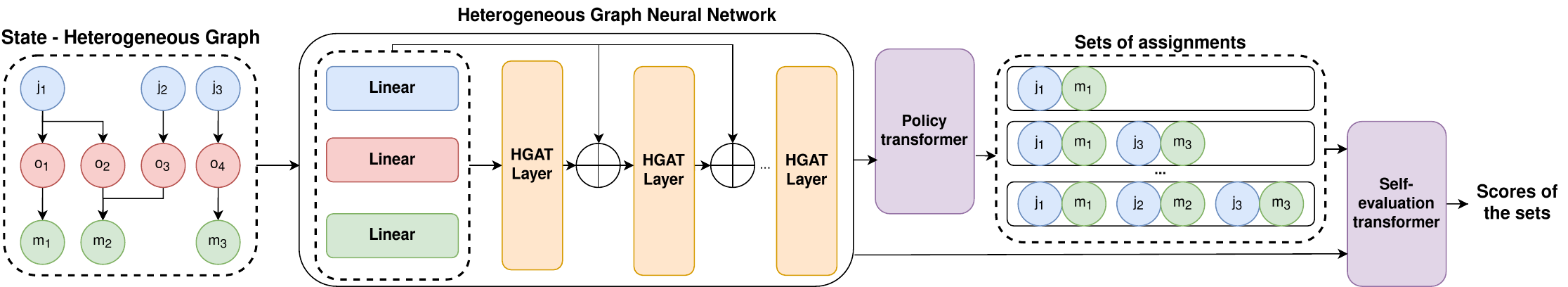} \caption{The architecture of the self-evaluation framework.} \label{fig:arcapp} \end{figure*}
 
\subsubsection{Policy Model Training}

The policy model predicts the probability distribution $\pi_\theta(a \mid s)$ over feasible job-machine assignments $a \in \mathcal{JM}$ for a given state $s$. To compute assignment probabilities, the model processes the state $s$ in two stages. First, a HGNN generates embeddings for the graph’s nodes. The HGNN is parameterized by $\phi$, uses $L$ propagation layers, and incorporates the attention mechanisms of GATv2 \cite{brody2021attentive} to effectively capture the structural and relational information of the state. Further details on the HGNN are provided in Appendix \ref{appendix:arquitecture}. Specifically, the HGNN computes embeddings $\boldsymbol{h_{j_{i}}^L}$ and $\boldsymbol{h_{m_{k}}^L}$, representing jobs and machines, respectively:
\begin{align}
\boldsymbol{h_{j_{i}}^L}, \boldsymbol{h_{m_{k}}^L} = \text{HGNN}_{\phi}(s).
\end{align}
Next, a Transformer processes these embeddings to compute the assignment probabilities. The input to the Transformer is a sequence that includes all feasible assignments $a \in \mathcal{JM}$. For each assignment, the input comprises the concatenation of the job embedding $\boldsymbol{h_{j_{i}}^L}$, the machine embedding $\boldsymbol{h_{m_{k}}^L}$, and the corresponding edge features $\boldsymbol{h_{mj_{ki}}}$. The Transformer then outputs the probability of the assignment:
\begin{align}
\mu_\theta(a | s) = \text{Transformer}_{\theta}([\boldsymbol{h_{j_{i}}^L},  \boldsymbol{h_{m_{k}}^L}, \boldsymbol{h_{mj_{ki}}]}).
\end{align}
The model is trained via supervised learning, minimizing the Kullback-Leibler (KL) divergence loss, as suggested in \cite{rusu2015policy}, to achieve more robust probability distributions. This loss measures the difference between the predicted distribution $\pi_\theta(a | s)$ and the target distribution $\pi_\text{solver}(a | s)$, which is derived from solver-generated optimal assignments:
\begin{align}
\mathcal{L}_{\text{policy}} = \text{KL}(\pi_\theta(a | s) , \pi_\text{solver}(a | s)).
\end{align}
Following this calculation, the weights of both the Transformer and the HGNN are updated to align the model’s predictions with the solver-provided target distribution.

\subsubsection{Self-Evaluation Model Training}

The self-evaluation model assigns a scalar score between 0 and 1 to subsets of job-machine assignments \( A \subseteq \mathcal{JM} \), representing the proportion of optimal assignments in the subset. In other words, a subset with no optimal assignments would receive a score of zero, while a subset where all assignments are optimal would receive a score of one. Once the model is trained, these scores are used during inference to  select high-quality sets of actions, ensuring that the generated solutions are closer to the optimal.

During training, the model leverages the embeddings $\boldsymbol{h_{j_{i}}^L}$ and $\boldsymbol{h_{m_{k}}^L}$ for jobs and machines, generated by the HGNN as part of the policy model training. Additionally, the embeddings for job-machine edges, $\boldsymbol{h_{mj_{ki}}}$, are included to encode relational information.

To mathematically represent sets of assignments, we model them as binary vectors. For each state \( s \) with its corresponding set of feasible job-machine assignments \( \mathcal{A} \subseteq \mathcal{JM} \), a binary vector \( \text{BV}(\mathcal{A}) \) is generated. This vector indicates whether each specific job-machine assignment is included in the subset, where a value of \( 1 \) denotes inclusion and \( 0 \) denotes exclusion.

During the training phase, for a given state \( s \) and its corresponding set of optimal assignments \( \mathcal{A}_{\text{opt}} \), a binary vector \( \text{BV}(A_\text{sub}) \) is randomly generated to simulate a subset of assignments. This vector indicates whether each job-machine pair is included in the subset and is compared against the optimal assignment vector to measure their similarity. The objective of the self-evaluation model is to learn to predict this similarity, which represents the degree to which a given subset resembles the optimal solution. This degree of similarity, referred to as the true score \( \text{TrueScore}(\mathcal{A}_\text{sub}),  \mathcal{A}_{\text{opt}} \), acts as the target during training and is computed using ground-truth information from the dataset.

To compute the predicted score for a given subset, the binary vector \( \text{BV}( \mathcal{A}_\text{sub}) \) is concatenated with the embeddings \( \boldsymbol{h_{j_{i}}^L} \), \( \boldsymbol{h_{m_{k}}^L} \), and \( \boldsymbol{h_{mj_{ki}}} \). These concatenated representations are processed by a Transformer architecture, which aggregates the information and applies mean pooling to produce a scalar score. The training process thus enables the self-evaluation model to assess the quality of any subset by estimating its proximity to the optimal solution.
\begin{align}
\text{SE}_\phi(A_\text{sub}) &= \text{MeanPooling} \Bigg( \text{Transformer}_{\phi}\bigg(     \nonumber \\
    &
        \big[\boldsymbol{h_{j_{i}}^L}, \boldsymbol{h_{m_{k}}^L}, \boldsymbol{h_{mj_{ki}}}, \text{BV}(A_\text{sub})\big]
    \bigg) 
\Bigg)
\end{align}
Figure~\ref{fig:arcapp} shows the overall architecture of the self-evaluation framework. The self-evaluation model is trained to minimize the Mean Squared Error (MSE) loss, aligning the predicted score $\text{SE}_\phi(A_\text{sub})$ with the true score $\text{TS}(A_\text{sub}, \mathcal{A}_{\text{opt}})$:
\begin{align}
\mathcal{L}_{\text{self-eval}} &= \frac{1}{|\mathcal{D}|} \sum_{|\mathcal{D}|} \big(\text{SE}_\phi(\mathcal{A}_\text{sub}) -  \nonumber \\
&\quad \text{TrueScore}(\mathcal{A}_\text{sub}, \mathcal{A}_{\text{opt}})\big)^2
\end{align}

In essence, the self-evaluation model aims to discern how closely a given subset resembles an optimal subset, providing critical feedback for selecting promising solutions during inference. The training process is summarized in Algorithm \ref{algo:training}, covering both the policy and self-evaluation models.

\begin{algorithm}[h]
\caption{Training of Policy and Self-Evaluation Models in SEVAL}
\label{algo:training}
\begin{algorithmic}[1]
\STATE \textbf{Input:} Dataset $\mathcal{D}$, parameters $\theta$ (policy), $\phi$ (self-evaluation), $\phi_{\text{HGNN}}$ (HGNN), epochs $n_e$.
\FOR{$epoch = 1$ to $n_e$}
    \FOR{each batch $(s, A_\text{opt}) \in \mathcal{D}$}
        \STATE \textbf{Step 1: Compute HGNN Embeddings}
        \[
        \boldsymbol{h_{j_{i}}^L}, \boldsymbol{h_{m_{k}}^L} = \text{HGNN}_{\phi_{\text{HGNN}}}(s)
        \]

        \STATE \textbf{Step 2: Train Policy Model}
        \STATE Generate assignment probabilities $a \in \mathcal{JM}_t$.
        \[
        \pi_\theta(a | s) = \text{Transformer}_{\theta}(\boldsymbol{h_{j_{i}}^L} || \boldsymbol{h_{m_{k}}^L} || \boldsymbol{h_{mj_{ki}}})
        \]
        \STATE Compute KL loss:
        \[
        \mathcal{L}_{\text{policy}} = \text{KL}(\pi_\theta(a | s), \pi_\text{solver}(a | s))
        \]
        \STATE Update $\theta$ and $\phi_{\text{HGNN}}$ using $\mathcal{L}_{\text{policy}}$.

        \STATE \textbf{Step 3: Train Self-Evaluation Model}
        \STATE Generate random subsets $A_\text{sub}$.
\begin{multline*}
\text{SE}_\phi(A_\text{sub}) = \text{MeanPooling} \big( \text{Transformer}_{\phi}\big(\\ 
\big[\boldsymbol{h_{j_{i}}^L}, \boldsymbol{h_{m_{k}}^L}, \boldsymbol{h_{mj_{ki}}}, \text{BV}(A_\text{sub})\big]\big)\big)
\end{multline*}
        \STATE Compute MSE loss:
\begin{multline*}
\mathcal{L}_{\text{self-eval}} = \frac{1}{|\mathcal{D}|} \sum_{|\mathcal{D}|} \big(\text{SE}_\phi(\mathcal{A}_\text{sub}) -   \\
\quad \text{TrueScore}(\mathcal{A}_\text{sub}, \mathcal{A}_{\text{opt}})\big)^2
\end{multline*}
        \STATE Update $\phi$ using $\mathcal{L}_{\text{self-eval}}$.
    \ENDFOR
\ENDFOR
\STATE \textbf{Output:} $\pi_\theta$, $\text{SE}_\phi$, $\phi_{\text{HGNN}}$
\end{algorithmic}
\end{algorithm}
During inference, the policy model and the self-evaluation model work together to determine the best subsets of actions at each timestep. At timestep $t$, the policy model generates $n$ candidate subsets by sampling from the learned probability distribution $\pi_\theta(a | s)$. Each subset is composed of $k$ feasible actions.\\

The self-evaluation model scores each of these subsets based on their optimality. The subset with the highest score is then selected for execution:
\begin{align}
A_t^* = \arg\max_{A \in \mathcal{A}_t} \text{SE}_\phi(A),
\end{align}
where $\mathcal{A}_t$ denotes the set of all candidate subsets generated by the policy model at timestep $t$, and $\text{SE}_\phi(A)$ is the scalar score assigned to each subset by the self-evaluation model.

\section{Experiments}
\label{sec:experiments}

In this section, we evaluate SEVAL, our proposed method. The code and implementation details will be made publicly available upon acceptance of the paper.

\subsection{Experimental Setup}

\textbf{JSSP Dataset.}  
The training dataset comprises 40,000 instances, with 90\% of the data used for training and 10\% for validation. All instances were solved using Google OR-Tools \cite{cpsatlp}, with a computation time limit of 60 seconds per instance. The data generation process followed the methodology proposed by Taillard \cite{taillard1993benchmarks}, generating instances with the number of jobs and machines ranging from 12 to 20 and processing times were uniformly sampled from the range $[1, 99]$.

\textbf{JSSP Test Benchmarks.}  
The performance of our method has been evaluated using two widely recognized benchmark datasets. The first is the Taillard dataset \cite{taillard1993benchmarks}, which contains 80 instances ranging from 15 jobs and 15 machines (225 operations) to 100 jobs and 20 machines (2000 operations). The second is the Demirkol dataset \cite{demirkol1998benchmarks}, which includes 80 instances with varying configurations, ranging from 20 to 50 jobs and machines. Notably, the Demirkol dataset introduces additional diversity by not following the same distribution as the Taillard dataset. The inclusion of both benchmarks allows us to evaluate the generalization capabilities of our approach on larger and more diverse instances than those used during training. The best-known results for these public benchmarks are available online\footnote{\url{https://optimizizer.com/index.php}}. 

\textbf{JSSP Baselines.}  
We compared our approach against several state-of-the-art methods. Among the DRL approaches, we included the actor-critic framework (L2D) proposed by \cite{zhang2020learning}, representing one of the pioneering works in this field, as well as the more recent HGNN-based approach, ResSch, introduced in \cite{ho2024residual}. 

From the methods allowing multiple assignments without employing self-evaluation mechanisms, we considered RLCP \cite{tassel2023end}, which uses a Transformer-based architecture and combines imitation learning with policy gradient methods. Additionally, we evaluated the approach proposed in \cite{echeverria2024multi}, which employs HGNNS and supervised learning and serves as the basis for this work. We also included emerging self-supervised learning approaches, such as SPN \cite{corsini2024self} and SI GD \cite{pirnay2024self}. For the comparison with SEVAL and other algorithms, the greedy strategy was chosen, selecting the action with the highest probability at each step, as our method generates a single solution without relying on sampling strategies.

For non-constructive methods, the improvement-based methods L2S \cite{zhang2024deep} was included, using its 500-step solution improvement variant to achieve comparable execution times. Finally, we compared our approach against the CP-SAT solver from Google OR-Tools, setting a time limit of 3600 seconds as a reference for a traditional optimization method.  
To ensure a fair comparison, we used the pre-trained models provided in the repositories of each paper to generate the results. Additionally, we compared execution times for all methods, which are presented in Appendix \ref{app:exec}.

\textbf{Performance Metric.}  
The evaluation metric used is the optimal gap ($OG$), defined as:
\begin{align}
OG = \left( \frac{C_{\pi}}{C_{\text{ub}}} - 1 \right) \cdot 100,
\end{align}
where $C_{\pi}$ is the makespan obtained by the policy, and $C_{\text{ub}}$ denotes the optimal or best-known makespan for the instance.

\textbf{Model Configuration.}  
The HGNN consisted of six layers, each with three attention heads and a hidden dimension of 32. The policy and self-evaluation models utilized a Transformer architecture with four layers, eight attention heads, latent and feed-forward dimensions of 128, and GeLU as the activation function. The number of subsets evaluated by the self-evaluation model was set to 16. The models were trained using the Adam optimizer with a learning rate of $3 \times 10^{-4}$ over 30 epochs and a batch size of 256.

\subsection{Experimental results}

\begin{table*}[!t]
\centering
\caption{Optimal gap comparison with several state-of-the-art DL methods, OR-Tools, and SEVAL in the Taillard benchmark. The best result for each group of instances among deep learning methods is highlighted in bold.}
\label{tab:allbenchmarkstaillard}
\begin{tabular*}{\textwidth}{l @{\extracolsep{\fill}} r r r r r r r r r}
\toprule
\multirow{1}{*}{\textbf{Size}} & \multicolumn{8}{c}{Deep learning methods} & \multicolumn{1}{c}{Classical-method} \\
\cmidrule(lr){2-9}
\cmidrule(lr){10-10}
& L2D & ResSch & RLCP & MAS &  SPN & SI GD & $L2S_{500}$ & \textbf{SEVAL} & OR-Tools \\
\midrule
15$\times$15 & 26.0 & 13.7 & 16.3 & 13.5 & 13.8 & 9.6 & 9.3 & \textbf{7.9} &  0.1 \\
20$\times$15 & 30.0 & 18.0 & 19.7 & 13.9 & 14.9 & 9.9& 11.6 & \textbf{7.6} & 0.2 \\
20$\times$20 & 31.6 & 16.5 & 18.6 & 13.4 & 15.2 & 11.1& 12.4 & \textbf{8.7} &  0.7\\
30$\times$15 & 33.0 & 17.3 & 18.3 & 14.8 & 17.1 & 9.5 & 14.7 & \textbf{9.6} &  2.1 \\
30$\times$20 & 33.6 & 18.1 & 22.8 & 17.4 & 18.5 & 13.8& 17.5 & \textbf{9.8} &  2.8 \\
50$\times$15 & 22.4 & 8.4 & 10.1 & 7.1 & 10.1 & \textbf{2.7} & 11.0 & 3.2 &  3.0\\
50$\times$20 & 26.5& 11.4 & 14.0 & 9.5 & 11.5 & 6.7& 13.0& \textbf{5.1} &  2.8\\
100$\times$20 & 13.6 & 4.0 & 4.5 & 2.3 & 5.8 & 1.7 & 7.9 & \textbf{0.5} &  3.9\\
Mean & 27.1 & 13.4 & 15.5 & 11.5 &  13.3 & 8.2 & 12.2 & \textbf{6.5} & 2.0\\
\bottomrule
\end{tabular*}
\end{table*}

\begin{table*}[!t]
\centering
\caption{Optimal gap comparison with several state-of-the-art DL methods, OR-Tools, and SEVAL in the Demirkol benchmark. The best result for each group of instances among deep learning methods, with similar execution times, is highlighted in bold.}
\label{tab:allbenchmarksdemirkol}
\begin{tabular*}{\textwidth}{l @{\extracolsep{\fill}} r r r r r r r r r}
\toprule
\multirow{1}{*}{\textbf{Size}} & \multicolumn{8}{c}{Deep learning methods} & \multicolumn{1}{c}{Classical-method} \\
\cmidrule(lr){2-9}
\cmidrule(lr){10-10}
& L2D & ResSch & RLCP & MAS & SPN & SI GD & $L2S_{500}$ & \textbf{SEVAL} & OR-Tools  \\
\midrule
20$\times$15 & 39.0& 19.6& 26.3 & 15.2 & 18.0& 15.4 & 21.1 & \textbf{11.8} & 1.8\\
20$\times$20 & 37.7& 18.9& 25.9 & 15.5 & 19.4 & 15.8 &18.2 & \textbf{10.7} & 1.9\\
30$\times$15 & 42.0& 20.6& 27.8 & 17.9 &21.8& 16.6 & 26.4 & \textbf{10.3} &2.5\\
30$\times$20 & 39.7& 22.1 & 30.1 & 19.1 &25.7& 16.1 & 27.5& \textbf{11.6} & 4.4\\
40$\times$15 & 35.6& 16.3& 22.6 & 13.3 &17.5& 11.6 & 25.8 & \textbf{7.5} & 4.1\\
40$\times$20 & 39.6& 19.2 & 27.8 & 18.4 &22.2& 16.6 & 29.2& \textbf{11.7} & 4.6\\
50$\times$15 & 36.5& 13.5 & 20.6 & 10.8 &15.7& 10.1 & 26.4& \textbf{6.2} & 3.8\\
50$\times$20 & 39.5& 20.1 & 26.4& 16.7 &22.4& 17.4 & 32.4 & \textbf{9.2} & 4.8\\
Mean & 38.7& 18.8 & 26.1 & 15.9 & 20.3 & 14.9 & 25.8 & \textbf{9.9} & 3.5\\
\bottomrule
\end{tabular*}
\end{table*}

\textbf{Results on the Taillard Benchmark.}  The experimental results on the Taillard benchmark, presented in Table \ref{tab:allbenchmarkstaillard}, show the outcomes for 80 instances grouped into sets of 10 by size. SEVAL, our proposed method, demonstrates superiority across all subsets except one. Among the deep learning methods, the best-performing results are highlighted, emphasizing SEVAL's consistent ability to outperform alternative approaches in most cases.

Notably, in the largest group of instances (\(100 \times 20\)), SEVAL achieves pseudo-optimal results with a mean gap of just \(0.5\%\), outperforming those obtained by OR-Tools, even when the latter is allocated an execution time of one hour per instance. As shown in Appendix \ref{app:exec}, SEVAL requires an average of only 30 seconds to solve these large instances.

Another noteworthy aspect is the remarkable improvement in performance metrics for this problem in recent years. For instance, the mean performance gap has decreased from $27.1\%$ in 2020 \cite{zhang2020learning} to $6.5\%$ in this work, suggesting that results very close to optimal may soon be achievable for this problem, much like the progress observed for simpler problems such as the TSP. Recent DRL-based approaches, such as ResSch \cite{ho2024residual}, have contributed to these improvements. However, methods based on supervised learning, like MAS \cite{echeverria2024multi} and SEVAL, or self-supervised learning, like SI GD \cite{pirnay2024self}, appear to achieve superior performance.

When comparing SEVAL to RLCP \cite{tassel2023end} and MAS \cite{echeverria2024multi}, both of which can assign multiple tasks and utilize a supervised learning component but lack a self-evaluation module, SEVAL demonstrates clear advantages. MAS, upon which this work builds, does not incorporate a hybrid HGNN architecture and entirely avoids the use of Transformers. These limitations may contribute significantly to the observed performance gap in favor of SEVAL.

SEVAL also achieves an improvement when compared to SI GD \cite{pirnay2024self}, the second-best performing method, which builds upon the work of \cite{drakulic2024bq} and employs a purely Transformer-based architecture. Similarly, it also outperforms the other self-supervised learning method, SPN \cite{corsini2024self}, which can also be attributed to its simpler architecture compared to the other methods.

Finally, SEVAL demonstrates competitive and superior results on smaller instances when compared to the L2S method proposed in \cite{zhang2024deep}. However, its performance advantage becomes more pronounced on larger instances. This difference can be attributed to the exponential growth in the search space and action complexity as instance size increases, which poses significant scalability challenges for improvement-based methods.

\textbf{Results on the Demirkol Benchmark.}  The results for the second benchmark, Demirkol, are presented in Table \ref{tab:allbenchmarksdemirkol}. The primary purpose of using this second benchmark, which has a distribution different from the one employed during training, is to evaluate the generalization capabilities of our method. 

The results are consistent with those observed on the Taillard benchmark, although the performance differences are more pronounced. Once again, SEVAL achieves the best results, with a significantly larger margin over the other methods compared to the Taillard benchmark. On the Taillard benchmark, the absolute difference between the second-best method and ours was \(1.7\%\), while on this benchmark, it is \(4\%\).

We also observe that while some degradation in performance is expected when transitioning to a different distribution, this drop is considerably higher for many methods. In some cases, the performance is halved, with the most severe degradation observed for the improvement-based method. In this case, the degradation is from \(12\%\) to \(25\%\), whereas in ours, it is from \(6.5\%\) to \(9.9\%\).

We hypothesize that this is due to the self-evaluation mechanism in SEVAL, which acts as an additional safeguard to the outputs generated by the policy. This mechanism contributes to producing more robust and reliable results, even when the data distribution shifts.

\section{Conclusion and Future Work}

In this paper, we propose a self-evaluation mechanism applied to the JSSP, achieving state-of-the-art performance on two widely recognized benchmarks. By evaluating subsets of actions collectively and leveraging a hybrid architecture of HGNNs and Transformers, our approach demonstrates strong generalization and robustness, outperforming classic methods like OR-Tools on larger instances. Future work includes combining supervised learning with reinforcement and self-supervised approaches, integrating beam search for improved inference, and extending the methodology to other combinatorial optimization problems, such as vehicle routing and resource allocation.









\bibliography{references}
\bibliographystyle{icml2025}

\newpage
\appendix
\onecolumn
\section{Features and edges of the state}
\label{app:features}

For job-type and operation-type nodes, the feature set is defined as:

\begin{itemize}
    \item \textbf{Binary Completion Indicator:} For jobs, $b_j \in \{0, 1\}$, where $b_j = 1$ if job $j$ is completed; otherwise, $b_j = 0$.
    \item \textbf{Operation Readiness Indicator:} For operations, $b_o \in \{0, 1\}$ indicates whether operation $o$ is ready to be scheduled.
    \item \textbf{Completion Time:} $t_j \in \mathbb{R}_{\geq 0}$ is the timestamp of the last completed operation for job $j$.
    \item \textbf{Mean Processing Time:} $p_o \in \mathbb{R}_{> 0}$ represents the expected duration of operation $o$.
    \item \textbf{Remaining Operations Count:} $r_j = |\mathcal{O}_j^{\text{rem}}|$ denotes the number of pending operations for job $j$.
    \item \textbf{Remaining Workload:} $s_j = \sum_{o \in \mathcal{O}_j^{\text{rem}}} \bar{p}_o$, where $\bar{p}_o \in \mathbb{R}_{> 0}$ is the processing time for operation $o$.
    \item \textbf{Job Completion Time Difference:} $t_j - t_{\text{min}}$, where $t_{\text{min}}$ is the minimum completion time across all jobs.
\end{itemize}

For machine-type nodes, the feature set is derived from machine-related metrics, capturing key information about machine states and performance:

\begin{itemize}
    \item \textbf{Number of Pending Operations:} $|\mathcal{O}_m^{\text{pending}}|$, the total number of operations yet to be assigned to machine $m$.
    \item \textbf{Assignable Operations Count:} $|\mathcal{O}_m^{\text{assignable}}|$, the number of first operations from each job that can currently be assigned to machine $m$.
    \item \textbf{Sum of Processing Times:} $\sum_{o \in \mathcal{O}_m^{\text{assignable}}} p_{o,m}$, the total processing time of all operations assignable to machine $m$.
    \item \textbf{Last Assigned Completion Time:} $t_{\text{final}}^m$, the completion time of the last operation assigned to machine $m$.
\end{itemize}

Edges in the graph capture relationships between nodes and are defined as:

\begin{itemize}
    \item \textbf{Undirected Edges (Machine-Operation):} Represent the compatibility between machines and operations, carrying features such as processing time.
    \item \textbf{Directed Edges (Operation-Job):} Define the relationship between operations and their jobs, indicating job ownership.
    \item \textbf{Directed Edges (Operation-Operation):} Enforce precedence constraints between dependent operations.
    \item \textbf{Directed Edges (Machine-Job):} Connect machines to the first pending operation of each job, including processing time features.
\end{itemize}

Two edge types carry specific features: operation-machine edges and job-machine edges. The features for operation-machine edges include:

\begin{itemize}
    \item \textbf{Processing Time:} $p_{o,m}$, the time required to execute operation $o$ on machine $m$.
    \item \textbf{Processing Time Relative to Job Workload:} $\frac{p_{o,m}}{\max_{o' \in \mathcal{O}_j} p_{o'}}$, where $\mathcal{O}_j$ is the set of remaining operations for job $j$, and $p_{o'}$ is the processing time of $o'$.
    \item \textbf{Processing Time Compared to Machine Capability:} $\frac{p_{o,m}}{\max{o' \in \mathcal{O}_m} p_{o'}}$, where $\mathcal{O}_m$ is the set of operations machine $m$ can process, and $p_{o'}$ is the processing time of $o'$.
\end{itemize}

\section{Example of Action Space Transition and State Transitions}
\label{app:example}

This section provides an illustrative example of the transitions in the action space and state representations for a JSSP instance. 


\begin{table}[h]
\caption{JSSP instance with 4 jobs and 4 machines where the processing times of each operation are indicated.}
\label{tab:jssp_example}
\begin{tabularx}{\columnwidth}{X X X X X X X}
\toprule
Jobs  & Operations & \(m_1\) & \(m_2\) & \(m_3\) & \(m_4\) \\ \midrule
\(j_1\) & \(o_{11}\) & - & - & - & 5 \\
       & \(o_{12}\) & - & 6 & - & - \\
       & \(o_{13}\) & 3 & - & - & - \\
       & \(o_{14}\) & - & - & 2 & - \\
\hline
\(j_2\) & \(o_{21}\) & - & - & - & 8 \\ 
       & \(o_{22}\) & 3 & - & - & - \\ 
\hline
\(j_3\) & \(o_{31}\) & - & - & 3 & - \\
       & \(o_{32}\) & 4 & - & - & - \\
       & \(o_{33}\) & - & - & - & 5 \\
\hline
\(j_4\) & \(o_{41}\) & - & 6 & - & - \\ 
       & \(o_{42}\) & - & - & - & 4 \\ 
       & \(o_{43}\) & - & - & 5 & - \\ \bottomrule
\end{tabularx}
\end{table}

The set of possible assignments in the initial step is given by:
\[
\mathcal{JM} = \{(j_1, m_4), (j_2, m_4), (j_3, m_3), (j_4, m_2)\}.
\]
The corresponding power set, ensuring that no machine is assigned more than one job simultaneously, defines the action space:
\begin{multline}
\mathcal{A} = \{ \{(j_1, m_4)\}, \{(j_2, m_4)\}, \dots, \{(j_1, m_4), (j_3, m_3)\}, \{(j_1, m_4), (j_4, m_2)\}, \dots, \\
\{(j_1, m_4), (j_3, m_3), (j_4, m_2)\} \}.
\end{multline}

An element of $\mathcal{A}$ is selected in each step, and the actions within this subset are executed. The corresponding operations are then removed from the graph, and the features of the remaining nodes are updated. This process repeats until all operations have been assigned.

As suggested in \cite{echeverria2024multi}, the action space of possible assignments is constrained to the maximum number of assignments (i.e., the number of machines), ordered by their minimum start times, to provide a sufficiently broad set of actions. This reduces the action space size, particularly for larger instances, enabling the policy to focus on a more manageable subset of options, which is especially advantageous in complex scenarios.


\section{Dataset Generation Process}
\label{app:dataset}

A diverse dataset is created by solving small-scale JSSP instances with an efficient solver. Solutions are converted into sequences of scheduling states, represented as heterogeneous graphs, along with their corresponding feasible job-machine assignments. Perturbing optimal solutions introduces variety, enhancing the model's generalization and performance in complex scenarios. To generate the dataset, the following steps were performed:

\begin{algorithm}[H]
\caption{Dataset Generation}
\begin{algorithmic}[1]
\STATE Initialize dataset $\mathcal{D} \gets \{\}$
\FOR{each JSSP instance $I_i \in \{I_1, \dots, I_n\}$}
    \STATE Solve $I_i$ to obtain the optimal trajectory and score: $\text{trajectory}_i = \{(s_t, \mathcal{A}_t)\}, \text{score}_\text{optimal}$
    \STATE Assign $n$ optimal actions, where $n \sim \text{Uniform}(0, 0.7 \cdot |I|)$
    \STATE Add $1$ to $30$ random actions to simulate perturbation
    \STATE Re-optimize perturbed assignments and calculate the new score: $\text{new\_score}_i$
    \IF{$\text{new\_score}_i / \text{score}_\text{optimal} > 1.1$}
        \STATE Ignore perturbed instance
    \ELSE
        \STATE Assign actions from the new trajectory
        \STATE Add to dataset: $\mathcal{D} \gets \mathcal{D} \cup \text{trajectory}_i$
    \ENDIF
\ENDFOR
\STATE Return $\mathcal{D}$
\end{algorithmic}
\end{algorithm}


\section{Model architecture}
\label{appendix:arquitecture}

In this section, we illustrate how embeddings are calculated within the HGNN model, using the machine embeddings as an example. The embedding process begins with a linear transformation applied to the initial features of machine, job, and operation nodes, denoted as $\boldsymbol{h_m}$, $\boldsymbol{h_o}$, and $\boldsymbol{h_j}$, respectively. These features represent the states of machines, operations, and jobs after the linear transformation, which ensures dimensional consistency. The embeddings are then iteratively updated across $L$ layers of the HGNN. At each layer, the new output is added to the input embedding, ensuring that the embeddings are progressively refined while retaining information from previous layers.

Machines aggregate information from connected operations. The attention coefficient between a machine $m_i$ and a connected operation $o_{ij} \in \mathcal{O}_{m_i}$ is computed as:
\begin{equation}
e_{mo_{ijk}} = {\boldsymbol{a^{\mathcal{MO}}}}^{\top} \text{LeakyReLU} \left( \boldsymbol{W_1^\mathcal{MO}} \boldsymbol{h_{m_i}} + \boldsymbol{W_2^\mathcal{MO}} \boldsymbol{h_{o_{ij}}} + \boldsymbol{W_3^\mathcal{MO}} \boldsymbol{h_{p_{ijk}}} \right),
\end{equation}
where ${\boldsymbol{a^{\mathcal{MO}}}}^{\top} \in \mathbb{R}^{3d_{\mathcal{MO}}^\prime}$ is a learnable vector, and $\boldsymbol{W_1^\mathcal{MO}}, \boldsymbol{W_2^\mathcal{MO}}, \boldsymbol{W_3^\mathcal{MO}} \in \mathbb{R}^{d_{\mathcal{MO}}^\prime \times d_{\mathcal{MO}}}$ are trainable weight matrices. The edge embedding $\boldsymbol{h_{p_{ijk}}}$ contains features of the task connecting the machine and the operation.

Similarly, machines also aggregate information from jobs if the first operation of a job $j$ is ready to be assigned to machine $m_i$. For these connections, the attention coefficient between a machine $m_i$ and a job $j \in \mathcal{J}$ is computed as:
\begin{equation}
e_{mj_{ik}} = {\boldsymbol{a^{\mathcal{MJ}}}}^{\top} \text{LeakyReLU} \left( \boldsymbol{W_1^\mathcal{MJ}} \boldsymbol{h_{m_i}} + \boldsymbol{W_2^\mathcal{MJ}} \boldsymbol{h_{j}} + \boldsymbol{W_3^\mathcal{MJ}} \boldsymbol{h_{mj_{ij}}} \right),
\end{equation}
where ${\boldsymbol{a^{\mathcal{MJ}}}}^{\top}$ and $\boldsymbol{W_1^\mathcal{MJ}}, \boldsymbol{W_2^\mathcal{MJ}}, \boldsymbol{W_3^\mathcal{MJ}}$ are learnable parameters, and $\boldsymbol{h_{mj_{ij}}}$ represents the features of the edge connecting the machine and job nodes. These edges are only considered if the first operation of the job is ready and can be assigned to the machine.

Once the attention coefficients are normalized using a softmax function, the updated embedding for a machine is calculated as:
\begin{equation}
\boldsymbol{h_{m_i}}^\prime = \text{ELU} \left( \sum_{o_{ij} \in \mathcal{O}_{m_i}} \alpha_{mo_{ijk}} \left( \boldsymbol{W_2^\mathcal{MO}} \boldsymbol{h_{o_{ij}}} + \boldsymbol{W_3^\mathcal{MO}} \boldsymbol{h_{p_{ijk}}} \right) + \sum_{j \in \mathcal{J}_{m_i}} \alpha_{mj_{ik}} \left( \boldsymbol{W_2^\mathcal{MJ}} \boldsymbol{h_{j}} + \boldsymbol{W_3^\mathcal{MJ}} \boldsymbol{h_{mj_{ij}}} \right) \right)
\end{equation}
where $\alpha_{mo_{ijk}}$ and $\alpha_{mj_{ik}}$ are the normalized attention coefficients for machine-operation and machine-job edges, respectively.

This process is repeated for $L$ layers of the HGNN. At each layer $l$, the input embedding $\boldsymbol{h_{m_i}}^{(l-1)}$ is updated as:
\begin{equation}
\boldsymbol{h_{m_i}}^{(l)} = \boldsymbol{h_{m_i}}^{(l-1)} + \boldsymbol{h_{m_i}}
\end{equation}
where $\boldsymbol{h_{m_i}}$ is the initial embedding.

\section{Execution time comparation}
\label{app:exec}

Tables \ref{tab:timeallbenchmarkstaillard} and Table \ref{tab:timeallbenchmarksdemirkol} present the computation times (in seconds) for deep learning methods applied to the Taillard and Demirkol benchmark problems. For all methods, the open-source implementations were utilized. It is important to note that execution time comparisons depend significantly on the specific implementation and the configuration of the computational environment.

In general, methods such as L2D and SPN exhibit faster execution times, primarily due to their simpler neural network architectures. Other methods require slightly longer computation times, which reflects the increased complexity of their architectures and algorithms. However, this additional complexity correlates with improved solution quality.

\begin{table*}[h]
\centering
\caption{Comparison of execution times (in seconds) between SEVAL and several state-of-the-art DL methods on the Taillard benchmark.}
\label{tab:timeallbenchmarkstaillard}
\begin{tabular*}{\textwidth}{l @{\extracolsep{\fill}} r r r r r r r r}
\toprule
\multirow{1}{*}{\textbf{Size}} & \multicolumn{8}{c}{Deep learning methods} \\
\cmidrule(lr){2-9}
& L2D & ResSch & RLCP & MAS & SPN & SI GD & L2S & \textbf{SEVAL} \\
\midrule
15$\times$15 & 0.7 &1.4 & 6.1 & 3.1 &0.5& 1.3 & 4.7 & 1.5  \\
20$\times$15 & 0.8 &1.3 & 6.8 & 3.7 &0.6 & 1.1 & 5.5 & 1.8  \\
20$\times$20 & 1.0 &2.8 & 7.55 & 5.0 &0.8 & 1.0 & 7.2 & 1.7 \\
30$\times$15 &1.1 &2.8 & 10.1 &5.4 & 0.9 & 1.1 & 8.5 &  2.1 \\
30$\times$20 &1.7 &3.3 & 11.6 &6.7 &1.3 & 1.5 & 10.2 &  2.5 \\
50$\times$15 & 2.3 &3.3 & 19.8 &12.1 &1.6 & 2.4 & 14.1 & 4.8 \\
50$\times$20 &3.4 &9.6 & 23.0 &13.8 &2.1 & 3.9 & 18.5 & 5.4\\
100$\times$20& 12.9& 9.8 &71.9 &58.2 &4.4 &30.1 & 47.2  & 41.5\\
Mean         &2.9&4.3 & 19.6 & 13.5 &1.5 & 5.1 & 14.5 & 7.6\\
\bottomrule
\end{tabular*}
\end{table*}

\begin{table*}[h]
\centering
\caption{Comparison of execution times (in seconds) between SEVAL and several state-of-the-art DL methods the Demirkol benchmark.}
\label{tab:timeallbenchmarksdemirkol}
\begin{tabular*}{\textwidth}{l @{\extracolsep{\fill}} r r r r r r r r}
\toprule
\multirow{1}{*}{\textbf{Size}} & \multicolumn{8}{c}{Deep learning methods} \\
\cmidrule(lr){2-9}
& L2D & ResSch & RLCP & MAS & SPN & SI GD & L2S & \textbf{SEVAL} \\
\midrule
20$\times$15 &0.7&1.1 & 6.8 &3.7 & 0.7& 1.1 & 9.0 & 1.4\\
20$\times$20 &0.9 &1.2 & 7.9 &4.7 &0.8 & 1.0 & 9.7 & 1.6 \\
30$\times$15 &1.1 &2.1 & 9.6 &5.2 &1.0 & 1.1 &  11.1 & 1.4\\
30$\times$20 &1.8 &2.5 & 11.8 &6.3 &1.2 & 1.5 &  14.1 & 1.9\\
40$\times$15 &2.8 &3.4 & 13.8 &7.4 &1.2 & 1.5 &  14.1& 2.0\\
40$\times$20 &2.7 &4.4 & 17.6 &8.5 &1.6 & 2.7 & 20.0  & 3.5\\
50$\times$15 &2.4 &5.0 & 19.1 &12.1 &1.5 &2.1 &  15.8 & 5.3\\
50$\times$20 &3.3 &6.2 & 28.2 &13.1 & 2.1& 3.9  & 18.1 & 5.7\\
Mean         &1.8 &3.2 & 14.3 &7.6  & 1.3& 1.8 & 13.9 & 2.8 \\
\bottomrule
\end{tabular*}
\end{table*}

\end{document}